\documentclass[10pt,twocolumn,letterpaper]{article}

\usepackage{cvpr}
\usepackage{times}
\usepackage{epsfig}
\usepackage{graphicx}
\usepackage{amsmath}
\usepackage{amssymb}
\usepackage{float}
\usepackage{caption}
\usepackage{subcaption}
\usepackage{booktabs}

\usepackage[pagebackref=true,breaklinks=true,letterpaper=true,colorlinks,bookmarks=false]{hyperref}

\cvprfinalcopy 

\def\cvprPaperID{721} 

\ifcvprfinal\fi
\begin{document}

\title{Viewpoints and Keypoints}

\author{Shubham Tulsiani and Jitendra Malik\\
University of California, Berkeley - Berkeley, CA 94720\\
{\tt\small \{shubhtuls,malik\}@eecs.berkeley.edu}
}

\maketitle
\thispagestyle{empty}

\begin{abstract}
We characterize the problem of pose estimation for rigid objects in terms of determining viewpoint to explain coarse pose and keypoint prediction to capture the finer details. We address both these tasks in two different settings - the constrained setting with known bounding boxes and the more challenging detection setting where the aim is to simultaneously detect and correctly estimate pose of objects. We present Convolutional Neural Network based architectures for these and demonstrate that leveraging viewpoint estimates can substantially improve local appearance based keypoint predictions. In addition to achieving significant improvements over state-of-the-art in the above tasks, we analyze the error modes and effect of object characteristics on performance to guide future efforts towards this goal.
\end{abstract}

\section{Introduction}

\begin{figure}[t!]
\centering
\includegraphics[width=0.45\textwidth]{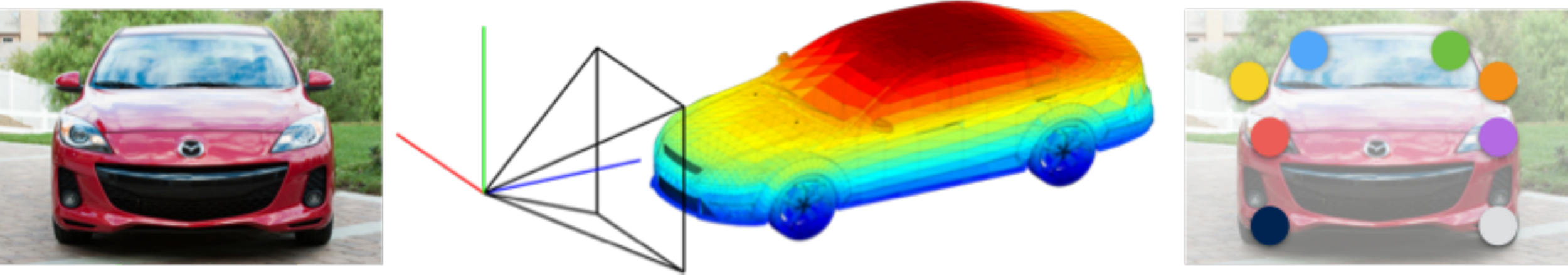}
\caption{Alternate characterizations of pose in terms of viewpoint and keypoint locations}
\label{figure:Fig1}
\end{figure}

There are two ways in which one can describe the pose of the car in Figure \ref{figure:Fig1} - either via its viewpoint or via specifying the locations of a fixed set of keypoints. The former characterization provides a global perspective about the object whereas the latter provides a more local one. In this work, we aim to reliably predict both these representations of pose for objects.

Our overall approach is motivated by the theory of global precedence - that humans perceive the global structure before the fine level local details \cite{globalPrecendence}. It was also noted by Koenderink and van Doorn \cite{koenderink1979internal} that viewpoint determines appearance and several works have shown that larger wholes improve the discrimination performance of parts \cite{configuralSuperiority,structuralPerception,palmer1981configural}.
Inspired by this philosophy, we propose an algorithm which first estimates viewpoint for the target object and leverages the predicted viewpoint to improve the local appearance based keypoint predictions.

Viewpoint is manifested in a 2D image by the spatial relationships among the different features of the object. Convolutional Neural Network (CNN) \cite{neocognitron,LeCun1989} based methods which can implicitly capture and hierarchically build on such relations are therefore suitable candidates for viewpoint prediction.

A robot which merely knows that a cup exists but cannot find its handle will not be able to grasp it. Towards the goal of developing a finer understanding of objects, we tackle the task of predicting keypoints by modeling appearances at multiple scales -  a fine scale appearance model, while prone to false positives can localize accurately and a coarser scale appearance model is more robust to mis-localizations. Note that merely reasoning over local appearance is not sufficient to solve the task of keypoint prediction. For example, the notion of the 'front wheel' assumes its meaning in context of the whole bicycle. The local appearance of the patch might also correspond to the 'back wheel' - it is because we know the bicycle is front facing that we are able to disambiguate. Motivated by this, we use the viewpoint predicted by our system to improve the local appearance based keypoint predictions.

\begin{figure*}[htb]
\centering
\includegraphics[scale=0.47]{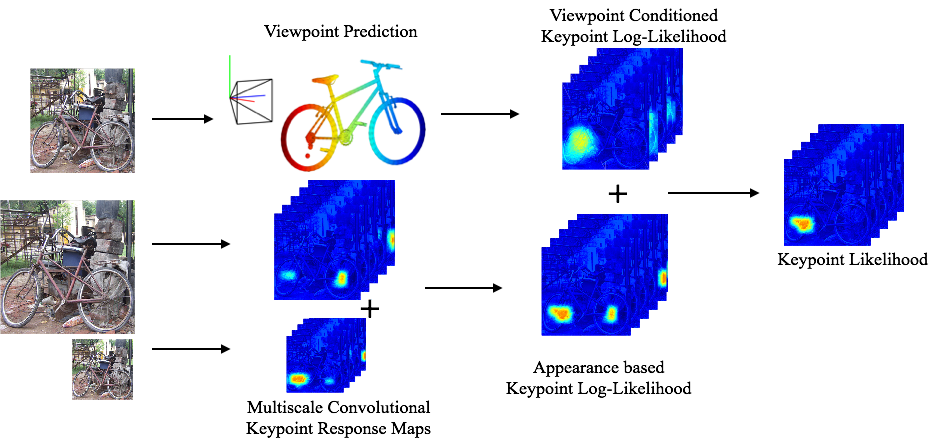}
\caption{Overview of our approach. To recover an estimate of the global pose, we use a CNN based architecture to predict viewpoint. For each keypoint, a spatial likelihood map is obtained via combining multiscale convolutional response maps and it is then combined with a likelihood conditioned on predicted viewpoint to obtain our final predictions.}
\label{figure:overviewFig}
\end{figure*}

Our proposed algorithm, as illustrated in Figure \ref{figure:overviewFig} has the following components -

\paragraph{Viewpoint Prediction :}
We formulate the problem of viewpoint prediction as predicting three euler angles ( azimuth, elevation and cyclorotation) corresponding to the instance. We train a CNN based architecture which can implicitly capture and aggregate local evidences for predicting the euler angles to obtain a viewpoint estimate.

\paragraph{Local Appearance based Keypoint Activation :}
We propose a fully convolutional CNN based architecture to model local part appearance. We capture the appearance at multiple scales and combine the CNN responses across scales to obtain a resulting heatmap  which corresponds to a spatial log-likelihood distribution for each keypoint.

\paragraph{Viewpoint Conditioned Keypoint Likelihood :}
We propose a viewpoint conditioned keypoint likelihood, implemented as a non-parametric mixture of gaussians, to model the probability distribution of keypoints given the viewpoint prediction. We combine it with the appearance based likelihood computed above to obtain our keypoint predictions.


Keypoint prediction methods have traditionally been evaluated assuming ground-truth boxes as input \cite{mpiiPose,LSP,JonNingNIPS}. This means that the evaluation setting is quite different from the conditions under which these methods would be used - in conjunction with imprecisely localized object detections. Yang and Ramanan \cite{YangRamanan} argued for the importance of this task for human pose estimation and introduced an evaluation criterion which we adapt to generic object categories. To the best of our knowledge, we are the first to empirically evaluate the applicability of a keypoint prediction algorithm not restricted to a specific object category in this challenging setting.

Furthermore, inspired by the analysis of the detection methods presented by Hoeim \etal \cite{hoiem2012diagnosing}, we present an analysis of our algorithm's failure modes as well as the impact of object characteristics on the algorithm's performance.



\section{Related Work}

\paragraph{Viewpoint Prediction: }
Recently, CNNs \cite{neocognitron,LeCun1989} have been shown to outperform Deformable Part Model (DPM) \cite{felzens_latent_pami10} based methods for recognition tasks \cite{rcnn,donahue2013decaf,Krizhevsky}. Whereas DPMs explicitly model part appearances and their deformations, the CNN architecture allows such relations to be captured implicitly using a hierarchical convolutional structure.  Girshick \etal \cite{DPMsCNNs} argued that DPMs could also be thought as a specific instantiation of CNNs and therefore training an end-to-end CNN for the corresponding task should outperform a method which instead explicitly models part appearances and relations.

This result is particularly applicable to viewpoint estimation where the prominent approaches, from the initial instance based methods \cite{huttenlocher1990recognizing} to current state-of-the-art  \cite{pascal3d,pepik12dpm} explicitly model local appearances and aggregate evidence to infer viewpoint. Pepik \etal \cite{pepik12dpm} extend DPMs to 3D to model part appearances and rely on these to infer pose and Xiang \etal \cite{pascal3d} introduce a separate DPM component corresponding to each viewpoint. Ghodrati \etal \cite{ghodrati14viewpoint} differ from the explicit part-based methodology, using a fixed global descriptor to estimate viewpoint. We build on both these approaches by using a method which, while using a global descriptor, can implicitly capture part appearances.

\paragraph{Keypoint Prediction: }
Keypoint Prediction can be classified into two settings - a)  'Keypoint Localization' where the task is to find keypoints for objects with known bounding boxes and b) 'Keypoint Detection'  where bounding box is unknown.  This problem has been particularly well studied for humans - tracing back from classic model-based approaches for video \cite{Rourke1980,Hogg1983} to more recent pictorial structure based approaches \cite{YangRamanan} on challenging single image based real world datasets like LSP\cite{LSP} or MPII Human Pose \cite{mpiiPose}. Recently Toshev \etal \cite{DeepPose}  demonstrated that CNN based models can successfully be used for keypoint prediction for humans and Tompson \etal \cite{TompsonJLB14} significantly improved upon these results using a purely convolutional approach. These evaluations, however, are restricted to keypoint localizations. A more general task of keypoint detection without assuming ground truth box annotations was also recently introduced for humans by Yang and Ramanan \cite{YangRamanan} and Gkioxari \etal \cite{kposelets,poseactionrcnn} evaluated their keypoint prediction algorithm in this setting.

For generic object categories, annotations for keypoints on the challenging PASCAL VOC dataset \cite{pascal-voc-2012} were introduced by Bourdev \etal \cite{bourdevECCV10}. Though similar annotations or fitted CAD models have been successfully used to train better object detection systems \cite{strongDPM} as well as for simultaneous object detection and viewpoint estimation \cite{pepik12dpm}, the task of keypoint prediction has largely been unaddressed for generic object categories. Long \etal \cite{JonNingNIPS} recently evaluated keypoint localization results across all PASCAL categories but, to the best of our knowledge, the more general setting of keypoint detection for generic object categories has not yet been explored.

Previous works \cite{Savarese_ICCV2007_Multiview,glasner2011viewpoint,HejratiR_NIPS_2012} have also jointly tackled the problem of keypoint detection and pose estimation. While these are perhaps the closest to ours in terms of goals, they differ markedly in methodology - they explicitly aggregate local evidence for pose estimation and have either been restricted to a specific object category \cite{glasner2011viewpoint,HejratiR_NIPS_2012} or use instance model based matching \cite{Savarese_ICCV2007_Multiview}. Long \etal \cite{JonNingNIPS}, on the other hand share many commonalities with our methodology for the task of keypoint prediction - convolutional keypoint detections augmented with global priors to predict keypoints. However, we show that we can significantly improve their results by combining multiscale convolutional predictions from a trained CNN with a more principled, viewpoint estimation based global model. Both \cite{HejratiR_NIPS_2012,JonNingNIPS} only evaluate keypoint localization performance whereas we also evaluate our method in the setting of keypoint detection.

\section{Viewpoint Estimation}
\label{sec:viewpoints}

\subsection{Formulation}
We formulate the global pose estimation for rigid categories as predicting the viewpoint wrt to a canonical pose. This is equivalent to determining the three euler angles corresponding to azimuth ($\phi$), elevation($\varphi$) and cyclo-rotation($\psi$). We frame the task of predicting the euler angles as a classification problem where the classes $\{1,\ldots N_{\theta}\}$ correspond to $N_{\theta}$ disjoint angular bins. We note that the euler angles, and therefore every viewpoint, can be equivalently described by a rotation matrix. We will use the notion of viewpoints, euler angles and rotation matrices interchangeably.

\subsection{Network Architecture and Training}

\label{sec:globalTraining}

Let $N_c$ be the number of object classes, $N_a$ be number of angles to be predicted per instance. The number of output units per class is $N_{a}*N_{\theta}$ resulting in a total of $N_{c}*N_{a}*N_{\theta}$ outputs. We adopt an approach similar to Girshick \etal \cite{rcnn} and finetune a CNN model whose weights are initialized from a model pretrained on the Imagenet \cite{imagenet_cvpr09} classification task. We experimented with the architectures from Krizhevsky \etal \cite{Krizhevsky} (denoted as TNet) and Simonyan \etal \cite{Simonyan14c} (denoted as ONet). The architecture of our network is the same as the corresponding pre-trained network with an additional fully-connected layer having $N_{c}*N_{a}*N_{\theta}$ output units. We provide an alternate detailed visualization of the network architecture in the supplementary material.

Instead of training a separate CNN for each class, we implement a loss layer that selectively considers the $N_{a}*N_{\theta}$ outputs corresponding the class of the training instance and computes  a logistic loss for each of the angle predictions. This allows us to train a CNN which can jointly predict viewpoint for all classes, thus enabling learning a shared feature representation across all categories. We use the Caffe framework \cite{jia2014caffe} to train and extract features from the CNN described above. We augment the training data with jittered ground-truth bounding boxes that overlap with the annotated bounding box with IoU $>$ 0.7. Xiang \etal \cite{pascal3d} provide annotations for $(\phi,\varphi,\psi)$ corresponding to all the instances in the PASCAL VOC 2012 detection train, validation set as well as for ImageNet images. We use the PASCAL train set and the ImageNet annotations to train the network described above and use the PASCAL VOC 2012 validation set annotations to evaluate our performance. 

\section{Viewpoint Informed Keypoint Prediction}
\label{section:keypoints}
	
As we noted earlier, parts assume their meaning in context of the whole. Thus, in addition to local appearance, we should take into account the global context.  To operationalize this observation, we propose a two-component approach to keypoint prediction.

\subsection{Multiscale Convolutional Response Maps}

We use CNN based architectures to learn the appearance of keypoints across an object class. Using a fully convolutional architecture allows us to capture local appearance in a more hierarchical and robust way than HOG feature based models while still allowing for efficient inference by sharing computations across evaluations at different spatial locations in the same image.

Let $C$ denote the set of classes, $K_c$ denote the set of keypoints for class $c$ and $N_c = |K_c|$. The total number of keypoints $N_{kp}$ is therefore $\underset{c \in C}{\sum}N_c$. We train a fully convolutional network with an input size $(384 \times 384)$ such that the channels in its last layer correspond to the keypoints i.e. we use a loss which forces the channels in the last layer to only fire at positions which correspond to the locations of the respective keypoint. The CNN architecture we use has the convolutional layers from ONet followed by an additional convolution layer with the output size $12 \times 12 \times N_{kp}$ such that each channel of the output corresponds to a specific keypoint of a particular class.

The architecture enforces that the receptive field of an output unit in the location $(i,j)$ has a centre corresponding to $(32*i,32*j)$ in the input image. For each training instance with annotated keypoints with locations $\{(x_k,y_k) | k \in K_c\}$, we construct a target response map $T$ with $T(k_i,k_j,k)=1$ and zero otherwise (where  $(k_i,k_j)$ is the index of the unit with its receptive field's centre closest to the annotated keypoint). For each keypoint, this is similar to training with multiple classification examples per image centered at the repective fields of output units, akin to the formulation used for object detection by Szegedy \etal \cite{szegedy2013deep}. Similar to the details described in section \ref{sec:globalTraining}, we use a loss layer that only selects the channels corresponding to the instance class and implements a euclidean loss between the output and the target map, thus enabling us to jointly train a single network to predict keypoints for all classes. We train using the annotations from Bourdev \etal \cite{bourdevECCV10} and use ground truth and jittered boxes as training examples.

The above network captures the appearance of the keypoints at a particular scale. A coarser scale would be more robust to false positives as it captures more context but would not be able to localize well. In order to benefit from the predictions at a coarser level, without compromising  localization, we propose using a multiscale ensemble of networks. We therefore train another network with exactly the same architecture with a smaller input size $(192 \times 192)$ and a smaller output size $6 \times 6 \times N_{kp}$. We upsample the outputs of the smaller network and linearly combine them with the outputs of the larger network to get a spatial log-likelihood response map $L(\cdot,\cdot,k)$ for each keypoint $k$.

\subsection{Viewpoint Conditioned Keypoint Likelihood}
If we know that a particular car is left-facing, we'd expect its left wheels to be visible but not the right wheels. In addition to the ability to predict visibility, we'd also have a strong intuition about the approximate locations of the keypoints. If the problem setting was restricted to a particular instance, the exact locations of the keypoints could be inferred geometrically from the exact global pose. However, the two assumptions that would allow this approach do not hold true - we have to deal with different instances of the object category and our inferred global pose would only be approximate. To counter this, we propose a non-parametric solution - we would expect the keypoints of a given instance to lie at positions similar to other training instances whose global pose is close to the predicted global pose for the given instance. 

Let the training instances for class $c$ be denoted by  $\{ R^i, \{(x^i_k,y^i_k) | k \in K_c \}\}$ where $R_i$ is the rotation matrix  and $\{(x^i_k,y^i_k) | k \in K_c\}$ the annotated keypoints corresponding to the $i_{th}$ instance. Let $R$ be the predicted rotation matrix corresponding to which we want a prior for keypoint locations denoted by $P$ st $P(i,j,k)$ indicates the likelihood of keypoint $k$ being present at location $(i,j)$. Let $\Delta(R_1,R_2) = \frac{ \| log(R_1^TR_2)\|_F}{\sqrt{2}}$ denote the geodesic distance between rotation matrices $R_1,R_2$ and $N(R) = \{i| \Delta(R,R_i) < \frac{\pi}{6} \}$ represent the the training instances whose viewpoint is close to the predicted viewpoint. Our non-parametric global pose conditional likelihood ($P$) is defined as a mixture of gaussians and we combine it with the local appearance likelihood $(L)$ to get keypoint locations as follows -
\begin{eqnarray}
P(\cdot,\cdot,k) =\frac{1}{| N(R) |}\underset{i \in N(R)}{\sum} \mathcal{N}((x^i_k,y^i_k),\sigma I) \\
(x_{k},y_{k}) = \underset{y,x}{argmax} ~ log(P(x,y,k)) + L(x,y,k)
\end{eqnarray}
Note that all the coordinates above are normalized by warping the instance bounding box to a fixed size ($12 \times 12$) and we choose $\sigma = 2$.

\newcommand{\gtViewWidth}{0.15}
\newcommand{\gtViewFormat}{png}
\begin{figure*}[htb!]
\includegraphics[width=\gtViewWidth\textwidth]{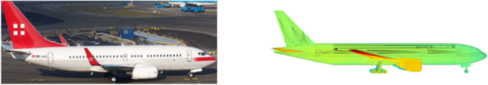} \hfill
\includegraphics[width=\gtViewWidth\textwidth]{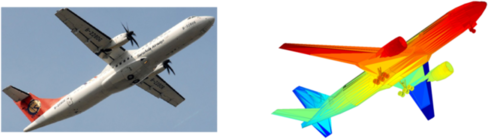}
\includegraphics[width=\gtViewWidth\textwidth]{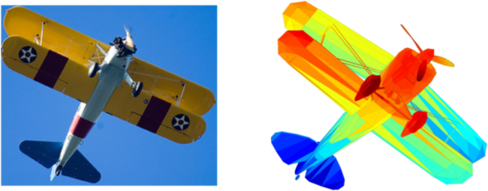} \hfill
\includegraphics[width=\gtViewWidth\textwidth]{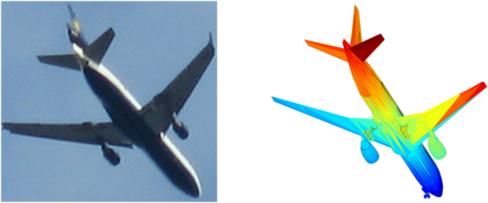} \hfill
\includegraphics[width=\gtViewWidth\textwidth]{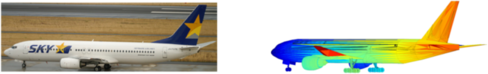}
\includegraphics[width=\gtViewWidth\textwidth]{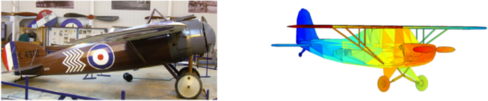}

\includegraphics[width=\gtViewWidth\textwidth]{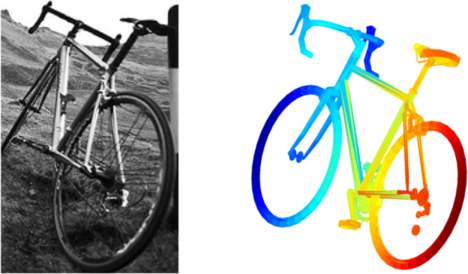} \hfill
\includegraphics[width=\gtViewWidth\textwidth]{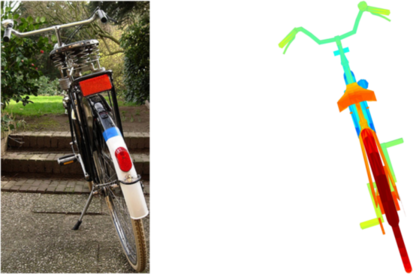} \hfill
\includegraphics[width=\gtViewWidth\textwidth]{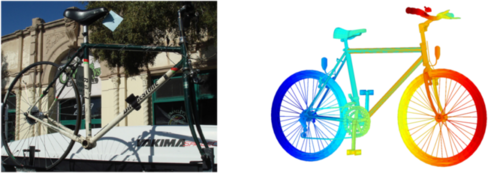}
\includegraphics[width=\gtViewWidth\textwidth]{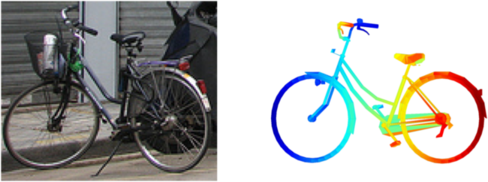} \hfill
\includegraphics[width=\gtViewWidth\textwidth]{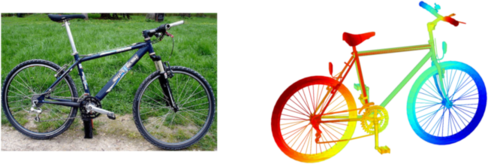} \hfill
\includegraphics[width=\gtViewWidth\textwidth]{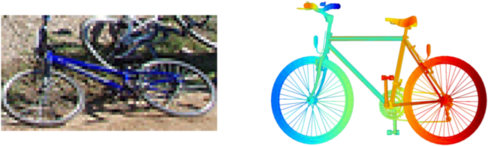}

\includegraphics[width=\gtViewWidth\textwidth]{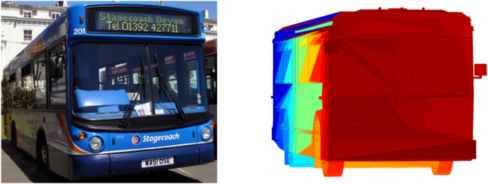} \hfill
\includegraphics[width=\gtViewWidth\textwidth]{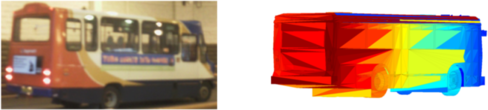} \hfill
\includegraphics[width=\gtViewWidth\textwidth]{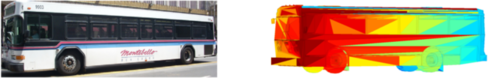}
\includegraphics[width=\gtViewWidth\textwidth]{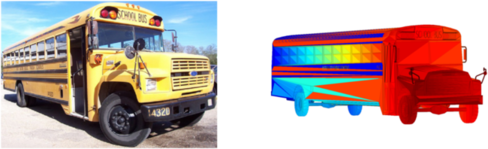} \hfill
\includegraphics[width=\gtViewWidth\textwidth]{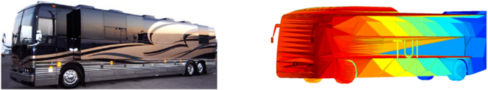} \hfill
\includegraphics[width=\gtViewWidth\textwidth]{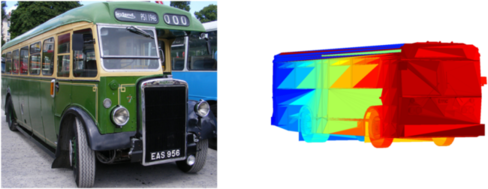}

\includegraphics[width=\gtViewWidth\textwidth]{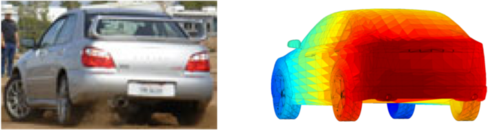} \hfill
\includegraphics[width=\gtViewWidth\textwidth]{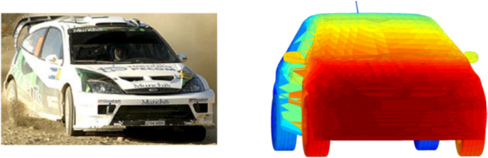} \hfill
\includegraphics[width=\gtViewWidth\textwidth]{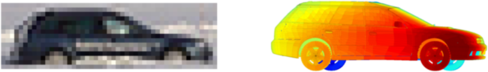}
\includegraphics[width=\gtViewWidth\textwidth]{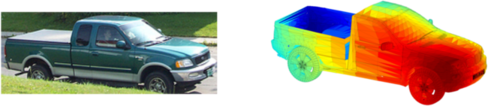} \hfill
\includegraphics[width=\gtViewWidth\textwidth]{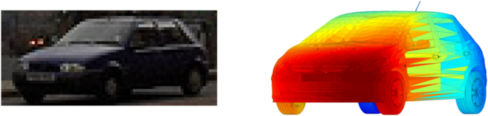} \hfill
\includegraphics[width=\gtViewWidth\textwidth]{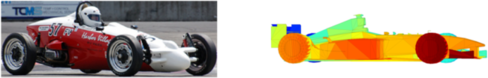}

\includegraphics[width=\gtViewWidth\textwidth]{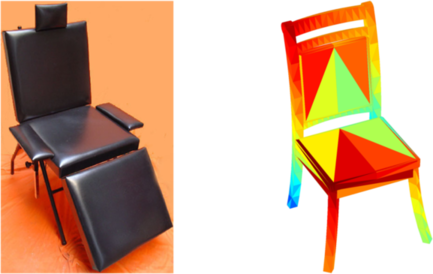} \hfill
\includegraphics[width=\gtViewWidth\textwidth]{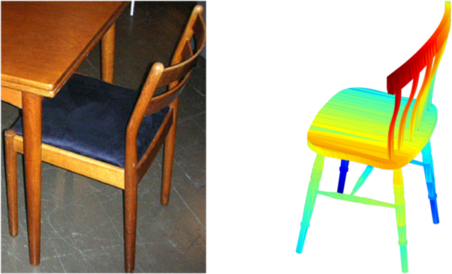} \hfill
\includegraphics[width=\gtViewWidth\textwidth]{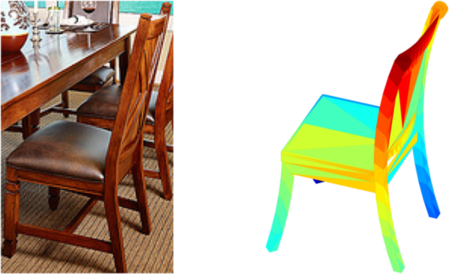}
\includegraphics[width=\gtViewWidth\textwidth]{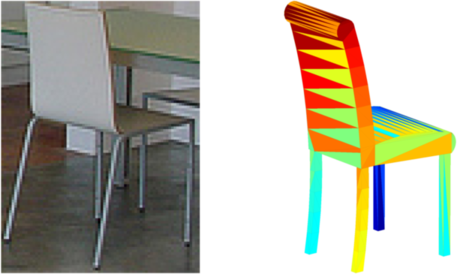} \hfill
\includegraphics[width=\gtViewWidth\textwidth]{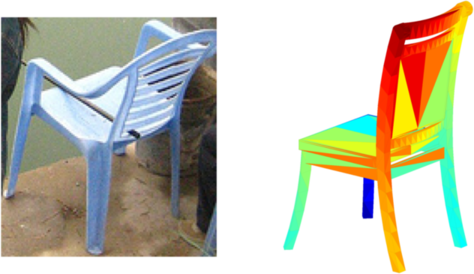} \hfill
\includegraphics[width=\gtViewWidth\textwidth]{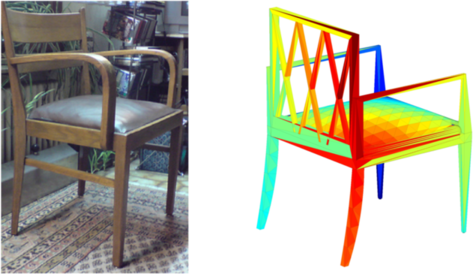}

\includegraphics[width=\gtViewWidth\textwidth]{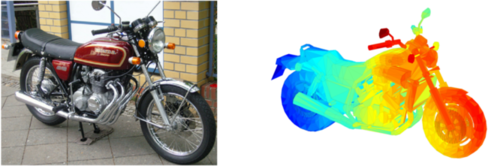} \hfill
\includegraphics[width=\gtViewWidth\textwidth]{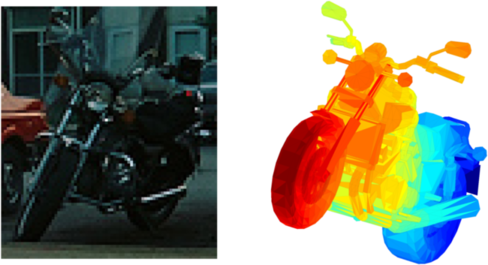} \hfill
\includegraphics[width=\gtViewWidth\textwidth]{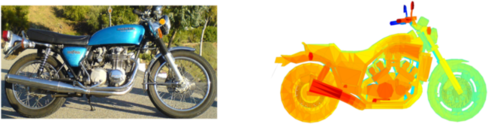}
\includegraphics[width=\gtViewWidth\textwidth]{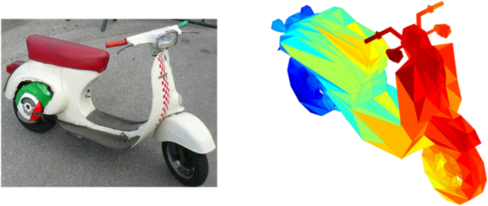} \hfill
\includegraphics[width=\gtViewWidth\textwidth]{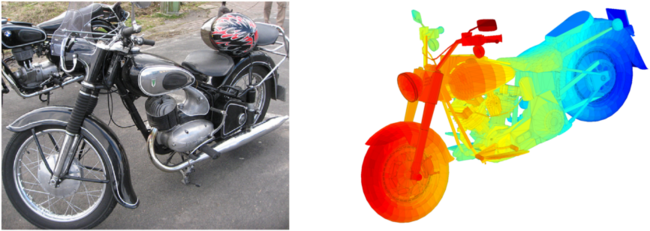} \hfill
\includegraphics[width=\gtViewWidth\textwidth]{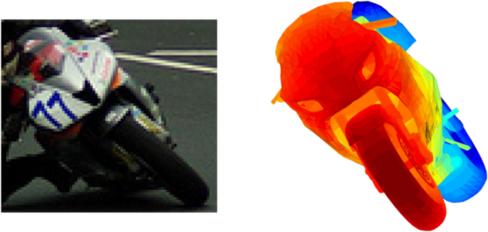}

\caption{Viewpoint predictions for unoccluded groundtruth instances using our algorithm.  The columns show 15th, 30th, 45th, 60th, 75th and 90th percentile instances respectively in terms of the error. We visualize the predictions by rendering a 3D model using our predicted viewpoint.}
\label{figure:viewpointPreds}
\end{figure*}

\section{Experiments : Viewpoint Prediction}

\begin{table*}[htb!]
\centering
\begin{tabular}{lcccccccccccc|c}
\toprule
 & \textbf{\footnotesize{}aero} & \textbf{\footnotesize{}bike} & \textbf{\footnotesize{}boat} & \textbf{\footnotesize{}bottle} & \textbf{\footnotesize{}bus} & \textbf{\footnotesize{}car} & \textbf{\footnotesize{}chair} & \textbf{\footnotesize{}table} & \textbf{\footnotesize{}mbike} & \textbf{\footnotesize{}sofa} & \textbf{\footnotesize{}train} & \textbf{\footnotesize{}tv} & \textbf{\footnotesize{}mean}\tabularnewline
\midrule
{\footnotesize{}$Acc_{\frac{\pi}{6}}$ (Pool5-TNet)} & {\footnotesize{}0.27} & {\footnotesize{}0.18} & {\footnotesize{}0.36} & {\footnotesize{}0.81} & {\footnotesize{}0.71} & {\footnotesize{}0.36} & {\footnotesize{}0.52} & {\footnotesize{}0.52} & {\footnotesize{}0.38} & {\footnotesize{}0.67} & {\footnotesize{}0.7} & {\footnotesize{}0.71} & {\footnotesize{}0.52}\tabularnewline
{\footnotesize{}$Acc_{\frac{\pi}{6}}$(fc7-TNet)} & {\footnotesize{}0.5} & {\footnotesize{}0.44} & {\footnotesize{}0.39} & {\footnotesize{}0.88} & {\footnotesize{}0.81} & {\footnotesize{}0.7} & {\footnotesize{}0.39} & {\footnotesize{}0.38} & {\footnotesize{}0.48} & {\footnotesize{}0.44} & {\footnotesize{}0.78} & {\footnotesize{}0.65} & {\footnotesize{}0.57}\tabularnewline
{\footnotesize{}$Acc_{\frac{\pi}{6}}$(ours-TNet)} & {\footnotesize{}0.78} & {\footnotesize{}0.74} & {\footnotesize{}0.49} & \textbf{\footnotesize{}0.93} & {\footnotesize{}0.94} & \textbf{\footnotesize{}0.90} & {\footnotesize{}0.65} & \textbf{\footnotesize{}0.67} & {\footnotesize{}0.83} & {\footnotesize{}0.67} & {\footnotesize{}0.79} & {\footnotesize{}0.76} & {\footnotesize{}0.76}\tabularnewline
{\footnotesize{}$Acc_{\frac{\pi}{6}}$(ours-ONet)} & \textbf{\footnotesize{}0.81} & \textbf{\footnotesize{}0.77} & \textbf{\footnotesize{}0.59} & \textbf{\footnotesize{}0.93} & \textbf{\footnotesize{}0.98} & {\footnotesize{}0.89} & \textbf{\footnotesize{}0.80} & \textbf{\footnotesize{}0}{\footnotesize{}.62} & \textbf{\footnotesize{}0.88} & \textbf{\footnotesize{}0.82} & \textbf{\footnotesize{}0.80} & \textbf{\footnotesize{}0.80} & \textbf{\footnotesize{}0.81}\tabularnewline
\midrule
{\footnotesize{}$MedErr$ (Pool5-TNet)} & {\footnotesize{}42.6} & {\footnotesize{}52.3} & {\footnotesize{}46.3} & {\footnotesize{}18.5} & {\footnotesize{}17.5} & {\footnotesize{}45.6} & {\footnotesize{}28.6} & {\footnotesize{}27.7} & {\footnotesize{}37} & {\footnotesize{}25.9} & {\footnotesize{}20.6} & {\footnotesize{}21.5} & {\footnotesize{}32}\tabularnewline
{\footnotesize{}$MedErr$(fc7-TNet)} & {\footnotesize{}29.8} & {\footnotesize{}40.3} & {\footnotesize{}49.5} & {\footnotesize{}13.5} & {\footnotesize{}7.6} & {\footnotesize{}13.6} & {\footnotesize{}45.5} & {\footnotesize{}38.7} & {\footnotesize{}31.4} & {\footnotesize{}38.5} & {\footnotesize{}9.9} & {\footnotesize{}22.6} & {\footnotesize{}28.4}\tabularnewline
{\footnotesize{}$MedErr$(ours-TNet)} & {\footnotesize{}14.7} & {\footnotesize{}18.6} & {\footnotesize{}31.2} & {\footnotesize{}13.5} & {\footnotesize{}6.3} & \textbf{\footnotesize{}8.8} & {\footnotesize{}17.7} & {\footnotesize{}17.4} & {\footnotesize{}17.6} & {\footnotesize{}15.1} & {\footnotesize{}8.9} & {\footnotesize{}17.8} & {\footnotesize{}15.6}\tabularnewline
{\footnotesize{}$MedErr$(ours-ONet)} & \textbf{\footnotesize{}13.8} & \textbf{\footnotesize{}17.7} & \textbf{\footnotesize{}21.3} & \textbf{\footnotesize{}12.9} & \textbf{\footnotesize{}5.8} & {\footnotesize{}9.1} & \textbf{\footnotesize{}14.8} & \textbf{\footnotesize{}15.2} & \textbf{\footnotesize{}14.7} & \textbf{\footnotesize{}13.7} & \textbf{\footnotesize{}8.7} & \textbf{\footnotesize{}15.4} & \textbf{\footnotesize{}13.6}\tabularnewline
\bottomrule
\end{tabular}
\caption{Viewpoint Estimation with Ground Truth box}
\label{table:poseGtEval}
\end{table*}

In this section, we use the the PASCAL3D+ \cite{pascal3d} annotations to evaluate the viewpoint estimation performance of our approach in the two different settings described below -

\subsection{Viewpoint Estimation with Ground Truth box}
\label{section:poseGtEval}
To analyze the performance of our viewpoint estimation method independent of factors like mis-localization, we first tackle the task of estimating the viewpoint of an object with known bounds. Let $\Delta(R_1,R_2) = \frac{ \| log(R_1^TR_2)\|_F}{\sqrt{2}}$ denote the geodesic distance function over the manifold of rotation matrices. $\Delta(R_{gt},R_{pred})$ captures the difference between ground truth viewpoint $R_{gt}$ and predicted viewpoint $R_{pred}$. We use two complementary metrics for evaluation -
\begin{itemize}
\item \textbf{Median Error :} The common confusions for the task of viewpoint estimation often are predictions which are far apart (eg. left facing vs right facing car) and the median error ($MedErr$) is a widely use metric that is robust to these if a significant fraction of the estimates are accurate.
\item \textbf{Accuracy at $\theta$ :} A small median error does not necessarily imply accurate estimates for all instances, a complementary performance measure is the fraction of instances whose predicted viewpoint is within a fixed threshold of the target viewpoint. We denote this metric by $Acc_{\theta}$ where $\theta$ is the threshold. We use $\theta = \frac{\pi}{6}$.
\end{itemize}

Recently, Ghodrati \etal \cite{ghodrati14viewpoint} achieved results comparable to state-of-the art by using a linear classifier over layer 5 features of TNet. We denote this method as 'Pool5-TNet' and implement it as a baseline. To study the effect of end-to-end training of the CNN architecture, we use a linear classifier on top of the fc7 layer of TNet as another baseline (denoted as 'fc7-TNet' ). With the aim of  analyzing viewpoint estimation independently, the evaluations were restricted only to objects marked as non-occluded and non-truncated and we defer the study of the effects of occlusion/truncation in this setting to section \ref{section:vpAnalysis}. The performance of our method and comparisons to the baseline are shown in Table \ref{table:poseDetEval}. The results clearly demonstrate that end-to-end training improves results and that our method with the TNet architecture performs significantly better than the 'Pool5-TNet' method used in \cite{ghodrati14viewpoint}. We also observe a significant improvement by using the ONet architecture and only use this architecture for further experiments/analysis. In figure \ref{figure:viewpointPreds}, we show our predictions sorted in terms of the error and  it can be seen that the predictions for most categories are reliable even at the 90th percentile.

\subsection{Detection and Viewpoint Estimation}
\label{section:poseDetEval}

Xiang \etal \cite{pascal3d} introduced the $AVP$ metric to measure advances in the task of viewpoint estimation in the setting where localizations are not known a priori. The metric is similar to the $AP$ criterion used for PASCAL VOC detection except that each detection candidate has an associated viewpoint and the detection is labeled correct if it has a correct predicted viewpoint bin as well as a correct localization (bounding box IoU $>$ 0.5). Xiang \etal \cite{pascal3d} also compared to Pepik \etal \cite{pepik12dpm} on the AVP metric using various viewpoint bin sizes and Ghodrati \etal \cite{ghodrati14viewpoint} also showed comparable results on the metric. To evaluate our method, we obtain detections from RCNN \cite{rcnn} using MCG \cite{mcg2014} object proposals and augment them with a pose predicted using the corresponding detection's bounding box. We note that there are two issues with the $AVP$ metric - it only evaluates the prediction for the azimuth ($\phi$) angle and discretizes viewpoint instead of treating it continuously. Therefore, we also introduce two additional evaluation metrics which follow the IoU $>$ 0.5 criteria for localization but modify the criteria for assigning a viewpoint prediction to be correct as follows -
\begin{itemize}
\item $AVP_{\theta}$ : $\delta(\phi_{gt},\phi_{pred})<\theta$
\item $ARP_{\theta}$ :  $\Delta(R_{gt},R_{pred})<\theta$
\end{itemize}
Note that $ARP_{\theta}$ requires the prediction of all euler angles instead of just $\phi$ and therefore, is a stricter metric.

The performance of our CNN based approach for viewpoint prediction is shown in Table \ref{table:poseDetEval} and it can be seen that we significantly outperform the state-of-the-art methods across all categories. While it is not possible to compare our pose estimation performance independent of detection with DPM based methods like \cite{pascal3d,pepik12dpm}, an indirect comparison results from the analysis using ground truth boxes where we demonstrate that our pose estimation approach is an improvement over \cite{ghodrati14viewpoint} which in turn performs similar to \cite{pascal3d,pepik12dpm} while using similar detectors.

\begin{table}[htb!]
\centering
\begin{tabular}{lcccccc}
\toprule
 &  \multicolumn{4}{c}{\footnotesize{}$AVP$} &  {\footnotesize{}$AVP_{\frac{\pi}{6}}$} & {\footnotesize{}$ARP_{\frac{\pi}{6}}$} \tabularnewline
\midrule
\textbf{\footnotesize{}Number of bins} & \textbf{\footnotesize{}4 } & \textbf{\footnotesize{}8} & \textbf{\footnotesize{}16} & \textbf{\footnotesize{}24} & \textbf{\footnotesize{}-} & \textbf{\footnotesize{}-}\tabularnewline
\midrule
\textbf{\footnotesize{}Xiang \etal \cite{pascal3d}} & {\footnotesize{}19.5} & {\footnotesize{}18.7} & {\footnotesize{}15.6} & {\footnotesize{}12.1} & {\footnotesize{}-} & {\footnotesize{}-}\tabularnewline
\textbf{\footnotesize{}Pepik \etal \cite{pepik12dpm}} & {\footnotesize{}23.8} & {\footnotesize{}21.5} & {\footnotesize{}17.3} & {\footnotesize{}13.6} & {\footnotesize{}-} & {\footnotesize{}-}\tabularnewline
\textbf{\footnotesize{}Ghodrati \etal \cite{ghodrati14viewpoint}} & {\footnotesize{}24.1} & {\footnotesize{}22.3} & {\footnotesize{}17.3} & {\footnotesize{}13.7} & {\footnotesize{}-} & {\footnotesize{}-}\tabularnewline
\textbf{\footnotesize{}ours} & \textbf{\footnotesize{}49.1} & \textbf{\footnotesize{}44.5} & \textbf{\footnotesize{}36.0} & \textbf{\footnotesize{}31.1} & {\footnotesize{}50.7} & {\footnotesize{}46.5}\tabularnewline
\bottomrule
\end{tabular}

\caption{Mean performance of our approach for various metrics. We report the performance for individual classes with the supplementary material}
\label{table:poseDetEval}
\end{table}

\begin{figure*}[htb!]
\centering
\includegraphics[width=0.95\textwidth]{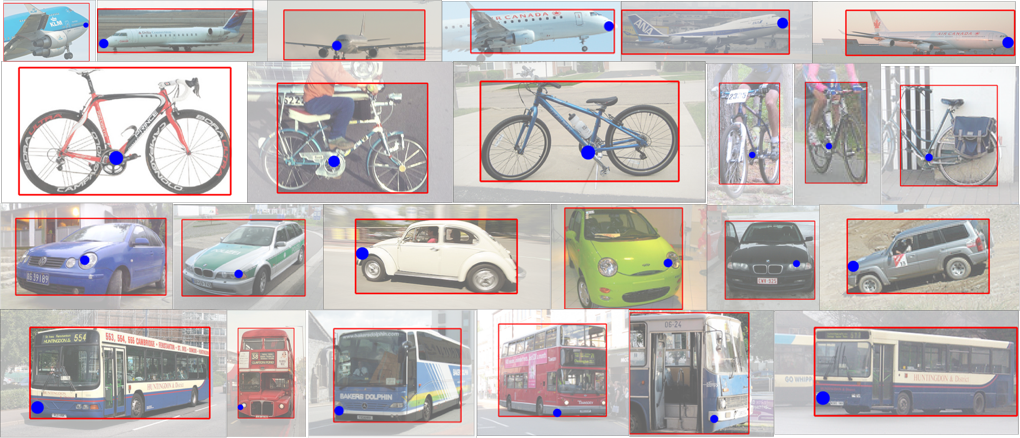}
\caption{Visualization of keypoints predicted in the detection setting. We visualize every $15^{th}$ detection, sorted by score, for 'Nosetip' of aeroplanes, 'Crankcentre' of bicycles, 'Left Headlight' of cars and 'Right Base' of buses.}
\label{figure:kpsDetsVis}
\end{figure*}

\section{Experiments : Keypoint Prediction}

\begin{center}
\begin{table*}[htb]
\centering
\begin{tabular}{lcccccccccccc|c}
\toprule
\textbf{\footnotesize{}PCK{[}$\alpha=0.1]$} & \textbf{\footnotesize{}aero} & \textbf{\footnotesize{}bike} & \textbf{\footnotesize{}boat} & \textbf{\footnotesize{}bottle} & \textbf{\footnotesize{}bus} & \textbf{\footnotesize{}car} & \textbf{\footnotesize{}chair} & \textbf{\footnotesize{}table} & \textbf{\footnotesize{}mbike} & \textbf{\footnotesize{}sofa} & \textbf{\footnotesize{}train} & \textbf{\footnotesize{}tv} & \textbf{\footnotesize{}mean}\tabularnewline
\midrule
\textbf{\footnotesize{}{Long \etal \cite{JonNingNIPS}}} & {\footnotesize{}53.7} & {\footnotesize{}60.9} & {\footnotesize{}33.8} & {\footnotesize{}72.9} & {\footnotesize{}70.4} & {\footnotesize{}55.7} & {\footnotesize{}18.5} & {\footnotesize{}22.9} & {\footnotesize{}52.9} & {\footnotesize{}38.3} & {\footnotesize{}53.3} & {\footnotesize{}49.2} & {\footnotesize{}48.5}\tabularnewline
\textbf{\footnotesize{}conv6 (coarse scale)} & {\footnotesize{}51.4} & {\footnotesize{}62.4} & {\footnotesize{}37.8} & {\footnotesize{}65.1} & {\footnotesize{}60.1} & {\footnotesize{}59.9} & {\footnotesize{}34.8} & {\footnotesize{}31.8} & {\footnotesize{}53.6} & {\footnotesize{}44} & {\footnotesize{}52.3} & {\footnotesize{}41.1} & {\footnotesize{}49.5}\tabularnewline
\textbf{\footnotesize{}conv12 (fine scale)} & {\footnotesize{}54.9} & {\footnotesize{}66.8} & {\footnotesize{}32.6} & {\footnotesize{}60.2} & {\footnotesize{}80.5} & {\footnotesize{}59.3} & {\footnotesize{}35.1} & {\footnotesize{}37.8} & {\footnotesize{}58} & {\footnotesize{}41.6} & {\footnotesize{}59.3} & {\footnotesize{}53.8} & {\footnotesize{}53.3}\tabularnewline
\textbf{\footnotesize{}conv6+conv12} & {\footnotesize{}61.9} & {\footnotesize{}74.6} & {\footnotesize{}43.6} & {\footnotesize{}72.8} & {\footnotesize{}84.3} & {\footnotesize{}70.0} & {\footnotesize{}45.0} & {\footnotesize{}44.8} & {\footnotesize{}66.7} & {\footnotesize{}51.2} & {\footnotesize{}66.8} & {\footnotesize{}56.8} & {\footnotesize{}61.5}\tabularnewline
\textbf{\footnotesize{}conv6+conv12+pLikelihood} & \textbf{\footnotesize{}66.0} & \textbf{\footnotesize{}77.8} & \textbf{\footnotesize{}52.1} & \textbf{\footnotesize{}83.8} & \textbf{\footnotesize{}88.7} & \textbf{\footnotesize{}81.3} & \textbf{\footnotesize{}65.0} & \textbf{\footnotesize{}47.3} & \textbf{\footnotesize{}68.3} & \textbf{\footnotesize{}58.8} & \textbf{\footnotesize{}72.0} & \textbf{\footnotesize{}65.1} & \textbf{\footnotesize{}68.8}\tabularnewline
\bottomrule
\end{tabular}
\caption{Keypoint Localization}
\label{table:kpsGtEval}
\end{table*}
\end{center}

\begin{table*}[htb]
\centering
\begin{tabular}{lcccccccccccc|c}
\toprule
\textbf{\footnotesize{}APK{[}$\alpha=0.1]$} & \textbf{\footnotesize{}aero} & \textbf{\footnotesize{}bike} & \textbf{\footnotesize{}boat} & \textbf{\footnotesize{}bottle} & \textbf{\footnotesize{}bus} & \textbf{\footnotesize{}car} & \textbf{\footnotesize{}chair} & \textbf{\footnotesize{}table} & \textbf{\footnotesize{}mbike} & \textbf{\footnotesize{}sofa} & \textbf{\footnotesize{}train} & \textbf{\footnotesize{}tv} & \textbf{\footnotesize{}mean}\tabularnewline
\midrule
\textbf{\footnotesize{}conv6+conv12} & {\footnotesize{}41.9} & {\footnotesize{}47.1} & {\footnotesize{}15.4} & {\footnotesize{}29.0} & {\footnotesize{}58.2} & {\footnotesize{}37.1} & {\footnotesize{}11.2} & {\footnotesize{}8.1} & {\footnotesize{}40.7} & {\footnotesize{}25.0} & {\footnotesize{}36.9} & {\footnotesize{}25.5} & {\footnotesize{}31.3}\tabularnewline
\textbf{\footnotesize{}conv6+conv12+pLikelihood} & \textbf{\footnotesize{}44.9} & \textbf{\footnotesize{}48.3} & \textbf{\footnotesize{}17.0} & \textbf{\footnotesize{}30.0} & \textbf{\footnotesize{}60.8} & \textbf{\footnotesize{}40.7} & \textbf{\footnotesize{}14.6} & \textbf{\footnotesize{}8.6} & \textbf{\footnotesize{}42.8} & \textbf{\footnotesize{}25.7} & \textbf{\footnotesize{}38.3} & \textbf{\footnotesize{}26.2} & \textbf{\footnotesize{}33.2}\tabularnewline
\bottomrule
\end{tabular}
\caption{Keypoint Detection}
\label{table:kpsDetEval}
\end{table*}

The task of keypoint prediction is commonly studied in the setting with known location of the object but some methods, restricted to specific categories like 'people' recently evaluated their performance in the more general detection setting. We extend these metrics to generic categories and evaluate our predictions in both the settings using the following metrics proposed by Yang and Ramanan \cite{YangRamanan} -

\begin{itemize}
\item PCK (Keypoint Localization) : For each annotated instance, the algorithm predicts a location for each keypoint and a groundtruth keypoint is said to have been found correctly if the corresponding prediction lies within $\alpha*max(h,w)$ of the annotated keypoint with the corresponding object's dimension being $(h,w)$. For each keypoint, we measure the fraction of objects where it was found correctly.
\item APK (Keypoint Detection) : A keypoint candidate is deemed correct if it lies within $\alpha*max(h,w)$ of a groundtruth keypoint. Each keypoint hypothesis has an associated score and the area under the precision-recall curve is used as the evaluation criterion.
\end{itemize}
We use the keypoint annotations from \cite{bourdevECCV10} and use the PASCAL VOC train set for training and the validation set images for evaluation.

\subsection{Keypoint Localization}
\label{sec:kpsGtEval}
The performance of our system and comparison to \cite{JonNingNIPS} are shown in Table \ref{table:kpsGtEval}. We denote by 'conv6' ('conv12') the predictions using only the $6 \times 6$  ($12 \times 12$) output size network, by 'conv6+conv12' the predictions using the multiscale convolutional response and by 'conv6+conv12+pLikelihood' the predictions using our full system. Our baseline system ( 'conv6+conv12') performs much better than  \cite{JonNingNIPS}, indicating the importance of end-to-end training and multiscale response maps. We also see that incorporating the viewpoint conditioned likelihood induces a significant performance gain.

\subsection{Keypoint Detection}
\label{sec:kpsDetEval}

Given an image, we use RCNN \cite{rcnn} combined with MCG \cite{mcg2014} object proposals to obtain detection candidates, each comprising of a class label and location. We then predict keypoints on each candidate using our system and score each keypoint hypothesis by linearly combining the keypoint log-likelihood score and the object detection system score. Our results for the task of keypoint detection are summarized in Table \ref{table:kpsDetEval}. The pose conditioned likelihood consistently improves the local appearance based predictions. Though the task of keypoint detection on PASCAL VOC has not yet been analyzed for categories other than person, we believe our results of 33.2\% mean APK with a reasonably strict threshold indicate a promising start.

The above results support our three main assertions - a global prior obtained in the form of a viewpoint conditioned likelihood improves the local appearance based predictions, that end-to-end trained CNNs can effectively model part appearances and combining responses from multiple scales significantly improves performance.

\subsection{Generalization to Articulated Pose}
While the focus of our work is pose prediction for rigid objects, we note that our multiscale convolutional response based approach is also applicable for articulated pose estimation. To demonstrate this, we trained our convolutional response map system to detect keypoints for the category 'person' in PASCAL VOC 2012 and achieved an APK = 0.22 which is a significant improvement compared to the state-of-the-art method \cite{poseactionrcnn} which achieves APK = 0.15. We refer the reader to \cite{poseactionrcnn} for further details on the evaluation metrics for the task of articulated pose estimation.

\section{Analysis}

 \begin{table}[htb!]
        \centering
        \begin{tabular}{lcc}
\toprule
\textbf{\footnotesize{}Setting} & \textbf{\footnotesize{}Mean Error} & \textbf{\footnotesize{}Mean Accuracy}\tabularnewline
\midrule
{\footnotesize{}Default} & {\footnotesize{}13.5} & {\footnotesize{}0.81}\tabularnewline
{\footnotesize{}Small Objects} & {\footnotesize{}15.1} & {\footnotesize{}0.75}\tabularnewline
{\footnotesize{}Large Objects} & {\footnotesize{}12.7} & {\footnotesize{}0.87}\tabularnewline
{\footnotesize{}Occluded Objects} & {\footnotesize{}19.9} & {\footnotesize{}0.65}\tabularnewline
\bottomrule
\end{tabular}
        \caption{Object characteristics vs viewpoint prediction error}
        \label{table:vpObjectModes}
    \end{table}

 \begin{table}[htb!]
 \centering
\begin{tabular}{cc}
\toprule
\textbf{\footnotesize{}Setting} & \textbf{\footnotesize{}Accuracy}\tabularnewline
\midrule
{\footnotesize{}Error$<\frac{\pi}{9}$} & {\footnotesize{}83.7}\tabularnewline
{\footnotesize{}$\frac{\pi}{9}<$Error $<\frac{2\pi}{9}$} & {\footnotesize{}5.7}\tabularnewline
{\footnotesize{}Error$>\frac{\pi}{9}$ \& Error($\pi-flip$)$<\frac{\pi}{9}$} & {\footnotesize{}5.8}\tabularnewline
{\footnotesize{}Error$>\frac{\pi}{9}$ \& Error($z-ref$)$<\frac{\pi}{9}$} & {\footnotesize{}6.5}\tabularnewline
{\footnotesize{}Other} & {\footnotesize{}2.9}\tabularnewline
\bottomrule
\end{tabular}
		\caption{Analysis of error modes for viewpoint prediction}
		\label{table:vpErrorModes}        
 \end{table}

\begin{table*}[htb!]
\centering
\begin{tabular}{lcccccccccccc|c}
\toprule
\textbf{\footnotesize{}PCK$[\alpha=0.1]$} & \textbf{\footnotesize{}aero} & \textbf{\footnotesize{}bike} & \textbf{\footnotesize{}boat} & \textbf{\footnotesize{}bottle} & \textbf{\footnotesize{}bus} & \textbf{\footnotesize{}car} & \textbf{\footnotesize{}chair} & \textbf{\footnotesize{}table} & \textbf{\footnotesize{}mbike} & \textbf{\footnotesize{}sofa} & \textbf{\footnotesize{}train} & \textbf{\footnotesize{}tv} & \textbf{\footnotesize{}mean}\tabularnewline
\midrule
{\footnotesize{}Default} & {\footnotesize{}66.0} & {\footnotesize{}77.8} & {\footnotesize{}52.1} & {\footnotesize{}83.8} & {\footnotesize{}88.7} & {\footnotesize{}81.3} & {\footnotesize{}65.0} & {\footnotesize{}47.3} & {\footnotesize{}68.3} & {\footnotesize{}58.8} & {\footnotesize{}72.0} & {\footnotesize{}65.1} & {\footnotesize{}68.8}\tabularnewline
{\footnotesize{}Occluded Objects} & {\footnotesize{}55.2} & {\footnotesize{}56.6} & {\footnotesize{}38.7} & {\footnotesize{}68.8} & {\footnotesize{}64.4} & {\footnotesize{}62.8} & {\footnotesize{}48.1} & {\footnotesize{}40.5} & {\footnotesize{}53.1} & {\footnotesize{}59.6} & {\footnotesize{}68.6} & {\footnotesize{}47.3} & {\footnotesize{}55.3}\tabularnewline
{\footnotesize{}Small Objects} & {\footnotesize{}51.6} & {\footnotesize{}66.4} & {\footnotesize{}48.1} & {\footnotesize{}81.2} & {\footnotesize{}85} & {\footnotesize{}67.4} & {\footnotesize{}57.4} & {\footnotesize{}48.2} & {\footnotesize{}57.9} & {\footnotesize{}53.8} & {\footnotesize{}57.4} & {\footnotesize{}56.8} & {\footnotesize{}60.9}\tabularnewline
{\footnotesize{}Large Objects} & {\footnotesize{}74.6} & {\footnotesize{}87.4} & {\footnotesize{}57.2} & {\footnotesize{}86.3} & {\footnotesize{}90.9} & {\footnotesize{}90.6} & {\footnotesize{}65.1} & {\footnotesize{}37.7} & {\footnotesize{}76.1} & {\footnotesize{}68.5} & {\footnotesize{}74.1} & {\footnotesize{}65.3} & {\footnotesize{}72.8}\tabularnewline
\midrule
{\footnotesize{}left/right} & {\footnotesize{}71.1} & {\footnotesize{}80.2} & {\footnotesize{}53.4} & {\footnotesize{}84.4} & {\footnotesize{}90.9} & {\footnotesize{}84.1} & {\footnotesize{}74.7} & {\footnotesize{}49.2} & {\footnotesize{}69.8} & {\footnotesize{}63.4} & {\footnotesize{}75.0} & {\footnotesize{}68.2} & {\footnotesize{}72.0}\tabularnewline
{\footnotesize{}PCK$[\alpha=0.2]$} & {\footnotesize{}79.9} & {\footnotesize{}88.7} & {\footnotesize{}69.1} & {\footnotesize{}95.2} & {\footnotesize{}92} & {\footnotesize{}88.3} & {\footnotesize{}79.6} & {\footnotesize{}67.5} & {\footnotesize{}87.3} & {\footnotesize{}72.2} & {\footnotesize{}82.2} & {\footnotesize{}78.1} & {\footnotesize{}81.7}\tabularnewline
\bottomrule
\end{tabular}
\caption{Analysis of Keypoint Prediction}
\label{table:kpAnalysis}
\end{table*}

An understanding of failure cases and effect of object characteristics on performance can often suggest insights for future directions. Hoeim \etal \cite{hoiem2012diagnosing} suggested some excellent diagnostics for object detection systems and we adapt those  for the task of pose estimation. We evaluate our system's output for both the task of viewpoint prediction as well as keypoint prediction but restrict our analysis to the setting with known bounding boxes - this enables us to analyze our pose estimation method independent of the detection system. We denote as 'large objects' the top third of instances and by 'small objects' the bottom third of instances. The label 'occluded' describes all the objects marked as truncated or occluded according to the PASCAL VOC annotations. We summarize our observations below.

\subsection{Viewpoint Prediction}
\label{section:vpAnalysis}
\paragraph{Object Characteristics : } Table \ref{table:vpObjectModes} shows the effect of object characteristics by reporting the mean across the classes of the median viewpoint error and accuracy. We can see that the method performs worse for occluded objects. There is also a significant difference between the performance for small and large objects - while such error trends are acceptable in the robotic setting where ambiguity for the farther objects is tolerable, one may need to capture more context to perform well without higher resolution input.

\paragraph{Error Modes: }
Since it is difficult to characterize error modes for generic rotations, we restrict the analysis to only the predicted azimuth. Assuming the image plane to be XY, we denote by $Z-ref$ the pose for the instance reflected along the XY plane and by $\pi-flip$ a rotation of $\pi$ along the $Z$ axis. Table \ref{table:vpErrorModes} reports the percentage of instances whose predicted pose corresponds to various modes. We observe that these error modes are equally common  and that only about $3\%$ of the errors are not explained by these.

Note that we exclude 'diningtable' and 'bottle' categories from the above analysis due to small number of unoccluded instances and insignificant variations respectively.

\subsection{Keypoint Prediction}

We use the PCK metric (section \ref{sec:kpsDetEval}) to characterize our algorithm's performance for various settings. Our results using the full method (local appearance combined with viewpoint conditioned likelihood) are reported in Table \ref{table:kpAnalysis}. We report the analysis using various components (single scale prediction, purely local appearance etc.) in the supplementary material.

\paragraph{Object Characteristics : }
The effect of object characteristics is similar to the viewpoint prediction setting - occluded objects are not handled well and there is a significant performance gap between small and large objects.

\paragraph{Error Modes : }
In the 'left/right' setting, we label a prediction to be correct if it was in the vicinity of the corresponding or the laterally inverted keypoint. Surprisingly, the performance is  similar to the base performance - indicating that laterally symmetric keypoints are not a significant error mode. The difference between the base performance and $PCK[\alpha=0.2]$ analyzes the inaccurate localizations which we find to be the main source of error.

\section{Conclusion}


We have presented an algorithm which leverages CNN architectures to predict viewpoint, and combines multiscale appearance with a viewpoint conditioned likelihood to predict keypoints. We demonstrated that our approach significantly improve state-of-the-art in settings with and without annotated bounding boxes for both viewpoint and keypoint prediction tasks. We also present evaluations for the keypoint detection setting alongwith a detailed ablation study of our performance on various tasks  and hope that these will contribute towards progress on the task of pose estimation for generic objects. We will make our code and trained models publicly available.

\section*{Acknowledgements}
The authors would like to thank Abhishek Kar, Jo\~{a}o Carreira and Saurabh Gupta for their valuable comments. This work was supported in part by NSF Award IIS-1212798 and ONR MURI - N00014-10-1-0933 and the Berkeley Graduate Fellowship. We gratefully acknowledge NVIDIA corporation for the donation of Tesla GPUs for this research.

{\small
\bibliographystyle{ieee}
\bibliography{rigidPose}

\begin{thebibliography}{10}\itemsep=-1pt

\bibitem{mpiiPose}
M.~Andriluka, L.~Pishchulin, P.~Gehler, and B.~Schiele.
\newblock 2d human pose estimation: New benchmark and state of the art
  analysis.
\newblock In {\em IEEE Conference on Computer Vision and Pattern Recognition
  (CVPR)}, June 2014.

\bibitem{mcg2014}
P.~Arbel\'{a}ez, J.~Pont-Tuset, J.~Barron, F.~Marques, and J.~Malik.
\newblock Multiscale combinatorial grouping.
\newblock In {\em Computer Vision and Pattern Recognition}, 2014.

\bibitem{strongDPM}
H.~Azizpour and I.~Laptev.
\newblock Object detection using strongly-supervised deformable part models.
\newblock In {\em Proceedings of the 12th European Conference on Computer
  Vision - Volume Part I}, ECCV'12, pages 836--849, Berlin, Heidelberg, 2012.
  Springer-Verlag.

\bibitem{bourdevECCV10}
L.~Bourdev, S.~Maji, T.~Brox, and J.~Malik.
\newblock Detecting people using mutually consistent poselet activations.
\newblock In {\em European Conference on Computer Vision (ECCV)}, 2010.

\bibitem{imagenet_cvpr09}
J.~Deng, W.~Dong, R.~Socher, L.-J. Li, K.~Li, and L.~Fei-Fei.
\newblock Imagenet: A large-scale hierarchical image database.
\newblock In {\em CVPR}, 2009.

\bibitem{donahue2013decaf}
J.~Donahue, Y.~Jia, O.~Vinyals, J.~Hoffman, N.~Zhang, E.~Tzeng, and T.~Darrell.
\newblock Decaf: A deep convolutional activation feature for generic visual
  recognition.
\newblock {\em arXiv preprint arXiv:1310.1531}, 2013.

\bibitem{pascal-voc-2012}
M.~Everingham, L.~Van~Gool, C.~K.~I. Williams, J.~Winn, and A.~Zisserman.
\newblock The {PASCAL} {V}isual {O}bject {C}lasses {C}hallenge 2012 {(VOC2012)}
  {R}esults.
\newblock
  http://www.pascal-network.org/challenges/VOC/voc2012/workshop/index.html.

\bibitem{felzens_latent_pami10}
P.~F. Felzenszwalb, R.~B. Girshick, D.~McAllester, and D.~Ramanan.
\newblock Object detection with discriminatively trained part-based models.
\newblock {\em TPAMI}, 2010.

\bibitem{neocognitron}
K.~Fukushima.
\newblock {N}eocognitron: {A} self-organizing neural network model for a
  mechanism of pattern recognition unaffected by shift in position.
\newblock {\em Biological Cybernetics}, 36:193--202, 1980.

\bibitem{ghodrati14viewpoint}
A.~Ghodrati, M.~Pedersoli, and T.~Tuytelaars.
\newblock Is 2d information enough for viewpoint estimation?
\newblock In {\em Proceedings of the British Machine Vision Conference}. BMVA
  Press, 2014.

\bibitem{rcnn}
R.~Girshick, J.~Donahue, T.~Darrell, and J.~Malik.
\newblock Rich feature hierarchies for accurate object detection and semantic
  segmentation.
\newblock In {\em CVPR}, 2014.

\bibitem{DPMsCNNs}
R.~B. Girshick, F.~N. Iandola, T.~Darrell, and J.~Malik.
\newblock Deformable part models are convolutional neural networks.
\newblock {\em CoRR}, abs/1409.5403, 2014.

\bibitem{poseactionrcnn}
G.~Gkioxari, B.~Hariharan, R.~Girshick, and J.~Malik.
\newblock R-cnns for pose estimation and action detection.
\newblock {\em CoRR}, abs/1406.5212, 2014.

\bibitem{kposelets}
G.~Gkioxari, B.~Hariharan, R.~Girshick, and J.~Malik.
\newblock Using k-poselets for detecting people and localizing their keypoints.
\newblock In {\em Computer Vision and Pattern Recognition (CVPR)}, 2014.

\bibitem{glasner2011viewpoint}
D.~Glasner, M.~Galun, S.~Alpert, R.~Basri, and G.~Shakhnarovich.
\newblock Viewpoint-aware object detection and pose estimation.
\newblock In {\em IEEE International Conference on Computer Vision (ICCV)},
  pages 1275--1282, 2011.

\bibitem{HejratiR_NIPS_2012}
M.~Hejrati and D.~Ramanan.
\newblock Analyzing 3d objects in cluttered images.
\newblock In P.~Bartlett, F.~Pereira, C.~Burges, L.~Bottou, and K.~Weinberger,
  editors, {\em Advances in Neural Information Processing Systems 25}, pages
  602--610. 2012.

\bibitem{Hogg1983}
D.~Hogg.
\newblock Model-based vision: a program to see a walking person.
\newblock {\em Image and Vision Computing}, 1(1):5 -- 20, 1983.

\bibitem{hoiem2012diagnosing}
D.~Hoiem, Y.~Chodpathumwan, and Q.~Dai.
\newblock Diagnosing error in object detectors.
\newblock In {\em Computer Vision--ECCV 2012}, pages 340--353. Springer Berlin
  Heidelberg, 2012.

\bibitem{huttenlocher1990recognizing}
D.~P. Huttenlocher and S.~Ullman.
\newblock Recognizing solid objects by alignment with an image.
\newblock {\em International Journal of Computer Vision}, 5(2):195--212, 1990.

\bibitem{jia2014caffe}
Y.~Jia, E.~Shelhamer, J.~Donahue, S.~Karayev, J.~Long, R.~Girshick,
  S.~Guadarrama, and T.~Darrell.
\newblock Caffe: Convolutional architecture for fast feature embedding.
\newblock {\em arXiv preprint arXiv:1408.5093}, 2014.

\bibitem{LSP}
S.~Johnson and M.~Everingham.
\newblock Clustered pose and nonlinear appearance models for human pose
  estimation.
\newblock In {\em Proceedings of the British Machine Vision Conference}, 2010.
\newblock doi:10.5244/C.24.12.

\bibitem{koenderink1979internal}
J.~J. Koenderink and A.~J. van Doorn.
\newblock The internal representation of solid shape with respect to vision.
\newblock {\em Biological cybernetics}, 32(4):211--216, 1979.

\bibitem{Krizhevsky}
A.~Krizhevsky, I.~Sutskever, and G.~E. Hinton.
\newblock Imagenet classification with deep convolutional neural networks.
\newblock In {\em NIPS}, 2012.

\bibitem{LeCun1989}
Y.~LeCun, B.~Boser, J.~S. Denker, D.~Henderson, R.~E. Howard, W.~Hubbard, and
  L.~D. Jackel.
\newblock Backpropagation applied to handwritten zip code recognition.
\newblock {\em Neural Comput.}, 1(4):541--551, Dec. 1989.

\bibitem{JonNingNIPS}
J.~Long, N.~Zhang, and T.~Darrell.
\newblock Do convnets learn correspondence?
\newblock In {\em NIPS}, 2014.

\bibitem{structuralPerception}
J.~McClelland and J.~Miller.
\newblock Structural factors in figure perception.
\newblock {\em Perception \& Psychophysics}, 26(3):221--229, 1979.

\bibitem{globalPrecendence}
D.~Navon.
\newblock {Forest before trees: The precedence of global features in visual
  perception}.
\newblock 1977.

\bibitem{Rourke1980}
J.~O'Rourke and N.~Badler.
\newblock Model-based image analysis of human motion using constraint
  propagation.
\newblock {\em Pattern Analysis and Machine Intelligence, IEEE Transactions
  on}, PAMI-2(6):522--536, Nov 1980.

\bibitem{palmer1981configural}
S.~E. Palmer and N.~M. Bucher.
\newblock Configural effects in perceived pointing of ambiguous triangles.
\newblock {\em Journal of Experimental Psychology: Human Perception and
  Performance}, 7(1):88, 1981.

\bibitem{pepik12dpm}
B.~Pepik, M.~Stark, P.~Gehler, and B.~Schiele.
\newblock Teaching 3d geometry to deformable part models.
\newblock In {\em IEEE Conference on Computer Vision and Pattern Recognition
  (CVPR)}, June 2012.

\bibitem{configuralSuperiority}
J.~R. Pomerantz, L.~C. Sager, and R.~J. Stoever.
\newblock Perception of wholes and of their component parts: Some configural
  superiority effects.
\newblock {\em Journal of Experimental Psychology-human Perception and
  Performance}, 3:422--435, 1977.

\bibitem{Savarese_ICCV2007_Multiview}
S.~Savarese and L.~Fei-Fei.
\newblock 3d generic object categorization, localization and pose estimation.
\newblock In {\em IEEE International Conference on Computer Vision (ICCV)},
  2007.

\bibitem{Simonyan14c}
K.~Simonyan and A.~Zisserman.
\newblock Very deep convolutional networks for large-scale image recognition.
\newblock {\em CoRR}, abs/1409.1556, 2014.

\bibitem{szegedy2013deep}
C.~Szegedy, A.~Toshev, and D.~Erhan.
\newblock Deep neural networks for object detection.
\newblock In {\em Advances in Neural Information Processing Systems}, pages
  2553--2561, 2013.

\bibitem{TompsonJLB14}
J.~Tompson, A.~Jain, Y.~LeCun, and C.~Bregler.
\newblock Joint training of a convolutional network and a graphical model for
  human pose estimation.
\newblock {\em CoRR}, abs/1406.2984, 2014.

\bibitem{DeepPose}
A.~Toshev and C.~Szegedy.
\newblock Deeppose: Human pose estimation via deep neural networks.
\newblock In {\em 2014 {IEEE} Conference on Computer Vision and Pattern
  Recognition, {CVPR} 2014, Columbus, OH, USA, June 23-28, 2014}, pages
  1653--1660. {IEEE}, 2014.

\bibitem{pascal3d}
Y.~Xiang, R.~Mottaghi, and S.~Savarese.
\newblock Beyond pascal: A benchmark for 3d object detection in the wild.
\newblock In {\em WACV}, 2014.

\bibitem{YangRamanan}
Y.~Yang and D.~Ramanan.
\newblock Articulated pose estimation with flexible mixtures-of-parts.
\newblock In {\em Proceedings of the 2011 IEEE Conference on Computer Vision
  and Pattern Recognition}, CVPR '11, pages 1385--1392, Washington, DC, USA,
  2011. IEEE Computer Society.

\end{thebibliography}
}

\end{document}